\pdfoutput=1

\documentclass[11pt]{article}

\usepackage[preprint]{acl}

\usepackage{times}
\usepackage{latexsym}
\usepackage{algorithmic}
\usepackage{algorithm}
\usepackage{xcolor}

\usepackage[T1]{fontenc}

\usepackage[utf8]{inputenc}

\usepackage{microtype}

\usepackage{inconsolata}

\usepackage{graphicx}
\usepackage{microtype}
\usepackage{graphicx}
\usepackage{subfigure}
\usepackage{booktabs} 
\usepackage{amsthm,amsmath,amssymb}
\usepackage{mathrsfs}
\usepackage{multirow}
\usepackage{colortbl}
\usepackage{hyperref}
\usepackage{amsmath}
\usepackage{amssymb}
\usepackage{mathtools}
\usepackage{amsthm}
\usepackage[capitalize,noabbrev]{cleveref}
\definecolor{mycolor1}{RGB}{234, 249, 252} 

\title{ProDS: Preference-oriented Data Selection for Instruction Tuning}

\author{
 \textbf{Wenya Guo\textsuperscript{1}},
 \textbf{Zhengkun Zhang\textsuperscript{2}},
 \textbf{Xumeng Liu\textsuperscript{1}},
 \textbf{Ying Zhang\textsuperscript{1}},
\\
 \textbf{Ziyu Lu\textsuperscript{1}},
 \textbf{Haoze Zhu\textsuperscript{1}},
 \textbf{Xubo Liu\textsuperscript{1}},
 \textbf{Ruxue Yan \textsuperscript{1}},
\\
\\
 \textsuperscript{1}College of Computer Science, Nankai University, 
 \textsuperscript{2}Baidu Inc. 
\\
 \small{
   \textbf{Correspondence:} \href{mailto:email@domain}{\{wenyaguo, yingzhang\}@nankai.edu.cn, zhangzhengkun01@baidu.com}, 
 }
}

\begin{document}
\maketitle
\begin{abstract}
Instruction data selection aims to identify a high-quality subset from the training set that matches or exceeds the performance of the full dataset on target tasks.
Existing methods focus on the instruction-to-response mapping, but neglect the human preference for diverse responses.
In this paper, we propose \textbf{Pr}eference-\textbf{o}riented \textbf{D}ata \textbf{S}election method (\textbf{ProDS}) that scores training samples based on their alignment with
preferences observed in the target set.
%
%
%
Our key innovation lies in shifting the data selection criteria from merely estimating features for accurate response generation to explicitly aligning training samples with human preferences in target tasks.
Specifically, {direct preference optimization (DPO) is employed to estimate human preferences across diverse responses. 
Besides, a bidirectional preference synthesis strategy is designed to score training samples according to both positive preferences and negative preferences.}
Extensive experimental results demonstrate our superiority to existing task-agnostic and targeted methods. 
%
%
%

\end{abstract}

\section{Introduction}
\label{sec:introduction}
\begin{figure}
    \centering
    \includegraphics[width=\linewidth]{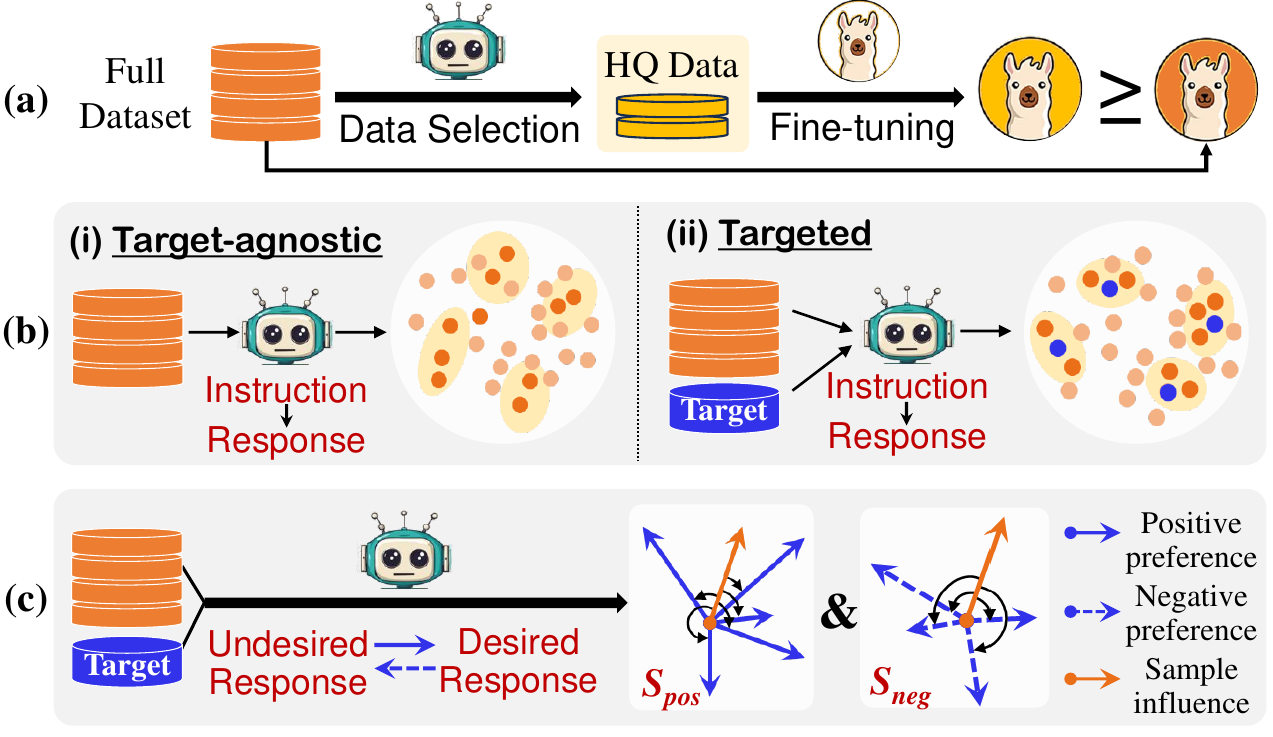}
    \caption{Illustration of (a) instruction data selection, (b) existing methods, and (c) our ProDS. 
    ``HQ Data'' denotes the selected high-quality instruction subset. 
    ``$S_{pos}$'' and ``$S_{neg}$'' are the consistency of training samples with the positive and negative preferences, respectively.
    }
    \label{fig:intro}
\end{figure}

Large language models (LLMs), such as GPT-3 \cite{GPT-3} and GPT-4 \cite{GPT-4}, have demonstrated outstanding capabilities in language understanding and generating.
%
Instruction tuning \cite{DBLP:conf/iclr/WeiBZGYLDDL22,longpre2023flan} is a crucial process during developing these notable capabilities by requiring LLMs to generate desired responses for specific instructions or guidelines. 
%
While early instruction tuning methods rely on vast datasets \cite{DBLP:conf/iclr/WeiBZGYLDDL22,wang-etal-2022-super}, \citeauthor{LIMA} highlight LIMA, which can achieve comparable performance to GPT-4 with only 1K carefully-constructed instruction-response pairs. 
LIMA inspires the research on representative instruction data selection. 
As shown in Fig.~\ref{fig:intro}(a), it aims to identify a high-quality subset (\textit{i.e.}, ``HQ Data'') from available datasets. 
LLMs fine-tuned on this subset can yield comparable or better performance to that using the full set.
%

Existing methods typically score the training samples based on the characteristics yielded in the process of instruction-based response generation.
%
According to whether introducing target tasks into data selection, they can be divided into target-agnostic methods \cite{IFD,MoDS,CaR,selectit} and targeted methods \cite{Nuggests,LESS}. 
%
As illustrated in Fig.~\ref{fig:intro}(b)-i, the target-agnostic approaches, such as IFD \cite{IFD} and MoDS \cite{MoDS}, score the training data according to the difficulty of generating responses, and select high-quality samples considering the coverage and diversity. 
%
In contrast, as shown in Fig.~\ref{fig:intro}(b)-ii, the targeted approaches, such as NUGGETS \cite{Nuggests} and LESS \cite{LESS}, directly associate the training samples with targeted tasks using instruction-to-response mapping information.
%
%

%


However, in practice, there is no absolute ``best'' response; there is always a more satisfactory one that ``better'' aligns with human preferences, {especially in the scenarios of open-domain Q\&A and dialogue.}
Correspondingly, for the target tasks, the adopted evaluators are not limited to verifying the consistency between generated responses and ground-truth references. 
They are also designed to assess whether a response is more preferred by humans compared to a baseline.
This preference for certain responses serves as a significant target-related indicator, as it reflects real-world user satisfaction for the specific target task. 
Yet, existing methods overlook this preference information, focusing instead on instruction-to-response mapping.
This raises a critical question: \textbf{
If responses aligned with human preferences are deemed superior, why not incorporate this preference information into the data selection process?}

In this paper, we propose \textbf{Pr}eference-\textbf{o}riented \textbf{D}ata \textbf{S}election method (\textbf{ProDS}), which aims to explicitly select training samples that align with the preference in target tasks.
%
\textit{Our main idea is to directly identify the samples that align with the preference for better responses}, rather than merely exploring the features for achieving the predefined best one.
%
%
To achieve this goal, we use direct preference optimization strategy (DPO) \cite{DPO} to
formulate preferences between various responses,
thus the gradient of DPO optimization can be regarded as {a representation of the preferences in target tasks.}
%
%

Besides, we design a \textbf{Bi}directional \textbf{P}reference \textbf{S}ynthesis strategy (\textbf{BiPS}) to utilize the preference features in two directions: 1) {a positive direction indicating an improvement in the quality of responses, shifting from undesired ones to desired ones};
 2) a negative direction signifying a decline in the quality of responses, transitioning from desired ones to undesired ones.
%
Ideally, high-quality samples should exhibit stronger alignment with positive directions and weaker alignment with negative ones.
%
%
As shown in Fig.~\ref{fig:intro}(c), we separately access the consistency of {the individual sample influence with both the positive and negative preferences.}
%
The positive and negative consistencies (\textit{i.e.}, ``$S_{pos}$'' and ``$S_{neg}$'') are then synthesized into the final scores for training samples using the annealing algorithm \cite{annealing}.
%

%

Our contributions can be summarized as:
1) 
We are the first to emphasize the response preference in instruction data selection. 
2) We design a bidirectional preference synthesis 
strategy to score training samples based on the correlation with positive and negative preferences.
%
3) Extensive experiments verify the superiority of our method to both target-agnostic and targeted data selection methods.

\section{Related Works}

\textbf{Instruction-tuning datasets.} 
%
Numerous human-annotated datasets are constructed to meet the demand on some specific tasks or more complex general instruction-following capabilities \cite{LIMA,Dolly}. 
However, the manual annotation process can inevitably introduce human bias \cite{CoachLM}. 
Thus, some other datasets are automatically generated using LLMs \cite{Selfinstruct,WizardLM}, which can provide a wealth of instances without human labor. 
The generated datasets contain more internal characteristics of LLMs and are also easier to expand and rewrite. 
Some pretrained models, such as Llama \cite{LLaMa} and Llama2 \cite{LLaMa2}, have shown excellent performance by fine-tuning on these datasets.

\textbf{Target-agnostic data selection methods.}
%
A main line of data selection methods is target-agnostic, relying on carefully designed assumptions about the characteristics of high-quality data, without considering the testing process. 
%
While early works use pre-defined features, such as some specific notions or n-grams \cite{xie2023data,chen2024skill}, 
recent methods focus more on the quality of input-output pairs and the diversity of them. 
For example, Li et al. \cite{Superfiltering,IFD} score instruction instances by evaluating the instruction-following difficulty.
%
And other methods such as MoDS \cite{MoDS} and CaR \cite{CaR} utilize cluster methods like K-Means \cite{Kmeans} to take data coverage and necessity into consideration.
%

\begin{figure*}
    \centering
    \includegraphics[width=.9\linewidth]{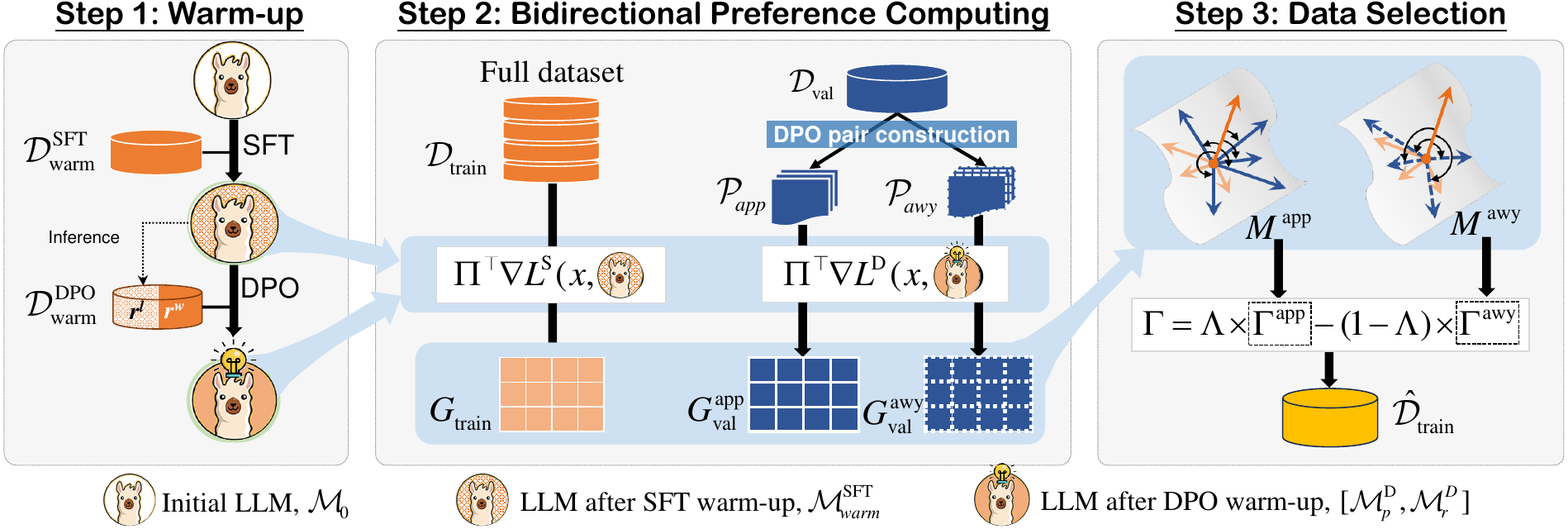}
    \caption{Illustration of our proposed ProDS.
    ProDS first uses the warmed-up models to compute bidirectional preference, and then scores the training samples by synthesizing their consistencies with the obtained preferences. 
    }
    \label{fig:pipeline}
\end{figure*}

\textbf{Targeted data selection methods.}
Another line of methods is targeted, directly incorporating the target-related information into the data selection process. 
Li et al. \cite{Nuggests} 
measure the impact of each training sample on the output of test data, and consider the samples that help reduce test loss as high-quality data.  
%
\citeauthor{han2023understanding} design an iterative gradient-based approach to search for 
instances which is beneficial for downstream in-context learning.
%
LESS \cite{LESS} employs a more straightforward method to select high-quality samples by evaluating the similarity between the response generation gradients of the training and target sets.
%
Compared to these methods, rather than focusing on the response generation process, we consider preferences for diverse responses and directly select samples aligned with the direction of improving response quality.

%
%

\section{Methodology}

For a full training dataset $\mathcal{D}_{\text{train}}$ with $N$ triplets $x=(Instruction, Input, Response)$, the task of instruction data selection aims to select a subset $\hat{\mathcal{D}}_\text{train}$ with high-quality and diverse triplets. 
%
The obtained  $\hat{\mathcal{D}}_\text{train}$ can be utilized to fine-tune a target LLM, $\mathcal{M}$, to equip the result LLM with satisfactory instruction-following capability.

Existing methods rely on the response generation process to identify high-quality and diverse triplets. 
%
In this paper, we propose a novel preference-oriented approach (ProDS) to identify triplets that align with human preferences.
ProDS directly incorporates response evaluation signals into the data selection process.
%
%
For a target dataset $\mathcal{D}_\text{val}$ with $L$ instructions, we first extract bidirectional preferences for diverse responses using the gradients from the direct preference optimization (DPO) process. 
%
Then, we employ an annealing-based scoring method to evaluate the similarity between training sample influence and obtained response preferences.
As shown in Fig.~\ref{fig:pipeline}, ProDS contains $3$ steps: base LLM warm-up, bidirectional preference computing, and data selection.
%
The details are stated as follows:

\subsection{Warm-up}
\label{sec:warm-up}
This phase contains two steps: a supervised fine-tuning (SFT) step and a direct preference optimization (DPO) step. 
Let $\mathcal{M}_0$ with parameters $\theta_0$ denote the initial pre-trained LLM. 
In the first step, we randomly select a small subset
from the training dataset, denoted as $\mathcal{D}_{\text{warm}}^{\text{SFT}} \subseteq \mathcal{D}_{\text{train}}$ to
fine-tune $\mathcal{M}_0$.
%
This fine-tuning process equips the result model, denoted as $\mathcal{M}_\text{warm}^\text{S}$ parameterized by $\theta^\text{S}$,  with the fundamental capability to follow instructions.
We use the cross-entropy loss for this SFT warm-up stage.
The SFT loss for the $N^\text{S}$ triplets in $\mathcal{D}_{\text{warm}}^{\text{SFT}}$ between the generated response (denoted as $r'_t$) and the ground-truth one (denoted as ${r}_t$) is:
\begin{equation}
    \mathcal{L}^\text{S} = \sum_{t=1}^{N^\text{S}} CELoss(r'_t, {r}_t).
\end{equation}

 In the second DPO warm-up step, we use $\mathcal{M}_\text{warm}^{\text{S}}$ to initialize the policy model $\mathcal{M}_{p}^{\text{D}}$ and reference model $\mathcal{M}_{r}^{\text{D}}$, where the parameters are denoted as $\theta^{\text{D}}_{p}$ and $\theta^{\text{D}}_{r}$, respectively. 
 And $\mathcal{M}_\text{warm}^{\text{SFT}}$ is also utilized to construct the warm-up preference dataset, 
 $\mathcal{D}_{\text{warm}}^{\text{DPO}}=\{i_t, r_t^w, r_t^l\}|_{t=1}^{N^\text{D}}$, where $i$ denotes the instructions and inputs, $r^w$ and $r^l$ denote the preferred responses and dispreferred responses, respectively.
 Specifically, $(i_t, r_t^w)$ is the input and response 
 in the original training dataset, $\mathcal{D}_{\text{train}}$, and $r_t^l$ indicates the output of $\mathcal{M}_\text{warm}^{\text{S}}$ for the same $i_t$.
With the constructed DPO warm-up dataset, we use the following loss function to further optimize the policy model $\mathcal{M}^\text{D}_p$:
%
\vspace{-5pt}
 \begin{equation}
 \footnotesize
\mathcal{L}^{\text{D}} =
  - \sum_{t=1}^{N^\text{D}}\log \sigma \left( \beta \log \frac{\mathcal{M}_p^\text{D}(r^w_t | i_t)}{\mathcal{M}_r^\text{D}(r^w_t | i_t)} 
  - \beta \log \frac{\mathcal{M}_p^\text{D}(r^l_t | i_t)}{\mathcal{M}_r^\text{D}(r^l_t | i_t)} \right),
  \label{equ:ld}
\end{equation}
where $\sigma(\cdot)$ is the sigmoid function and $\beta$ is a  temperature parameter for scaling implicit reward differences.
Since $\mathcal{M}_r^\text{D}$ is fixed during the DPO warm-up process, $\theta_p^{\text{D}}$ is the same as $\theta^\text{S}$.
%

\subsection{Bidirectional Preference Computing}
After the warm-up process, the obtained model is equipped with the capabilities of both following instructions and aligning with human preferences.
Drawing inspiration from \cite{LESS,zhang2024tagcos}, we utilize gradient to estimate the {the influence of training samples} and human preferences, based on which to associate the training data and the target tasks. 

{\textbf{Training sample influence computing. }}
For a triplet $x\in \mathcal{D}_\text{train}$, we can obtained the training gradient by $\nabla \mathcal{L}^\text{S}(x, \theta^\text{S}) \in \mathbb{R}^P$, where $P$ is the dimension of the gradient vector. 
{The influence of the individual sample $x$ is}:

\begin{equation}
    G_\text{train}=\{\Pi^\top \nabla \mathcal{L}^\text{S}(x, \theta^\text{S})\}|_{x\in \mathcal{D}_\text{train}} \in \mathbb{R}^{N\times d}.
\end{equation}
Following LESS \cite{LESS}, the used $\Pi^\top \in \mathbb{R}^{P\times b}$ is drowned from a Rademacher distribution (\textit{i.e.}, $\Pi_{ij}\sim U(\{-1,1\})$ to further improve memory efficiency \cite{park2023trak}.
In practice, we set $d=8192$.

\begin{figure}
    \centering
    \includegraphics[width=\linewidth]{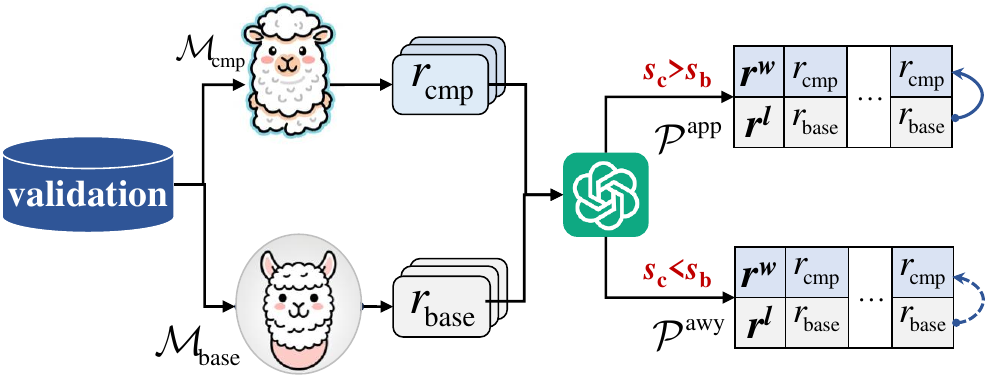}
    \caption{Illustration of the validation DPO pair construction process, where ``$s_c$'' and ``$s_b$'' are scores given by GPT-4 for the quality of ``$r_\text{cmp}$'' and ``$r_\text{base}$''.
    }
    \label{fig:dpoconstruct}
\end{figure}

\textbf{Bidirectional preference computing for target sets. }
Using the loss function in Equation~\eqref{equ:ld}, the result LLM has learned 
to assign higher rewards to preferred pairs, ($i_t$, $r^w_t$), than dispreferred ones, ($i_t$,$r^l_t$). 
The gradient indicates the direction to achieve this objective, reflecting human preference for superior responses. 
%

We first extract a subset from the target task, forming a validation set, $\mathcal{D}_\text{val}=\{i_{t}\}|_{t=1}^L$, where $i_t$ denotes the instruction and input.
Based on $\mathcal{D}_\text{val}$, we construct the corresponding DPO pairs. 
The process of DPO pair construction is shown in Fig.~\ref{fig:dpoconstruct}.
%
We use two different models, denoted as $\mathcal{M}_\text{base}$ and $\mathcal{M}_\text{cmp}$, to generate responses of varying quality for the same instruction, producing $r_\text{base}$ and $r_\text{cmp}$. 
GPT-4 is used to judge the quality of the obtained responses.
%
%
Building upon the intuition that high-quality data should approach correct preferences and keep away from incorrect ones,
we consider human preferences in two directions: one is positive preference that the response changes from undesired to desired, and the other is the opposite. 
Correspondingly, we construct two sets of DPO pairs: 
%
\begin{equation}
\begin{split}
\small
    & \mathcal{P}^\text{app}=\{i^t, r_w^t=r_\text{cmp}^t, r_l^t=r_\text{base}^t\}|_{t=1}^{L^\text{app}}, \text{if} ~ s_{c}^t > s_{b}^t,\\
    & \mathcal{P}^\text{awy}=\{i^t, r_w^t=r_\text{cmp}^t, r_l^t=r_\text{base}^t\}|_{t=1}^{L^\text{awy}}, \text{if}~s_{c}^t < s_{b}^t,
    \end{split}
\end{equation}
where $L^\text{app}+L^\text{awy}=L$, and $s_{c}^t$ and $s_{b}^t$ are the GPT-4 scores for $r_\text{cmp}^t$ and $r_\text{base}^t$, respectively.
%
Based on the DPO pairs, we can compute the corresponding bidirectional preference as follows:
\begin{equation}
\begin{split}
\small
  G_\text{val}^\text{app}=\{\Pi^\top \nabla & \mathcal{L}^\text{D}({x'}, \theta^\text{D}_p, \theta^\text{D}_{r})\}|_{x'\in \mathcal{P}^\text{app}} \in \mathbb{R}^{L^\text{app}\times d}, \\
  G_\text{val}^\text{awy}=\{\Pi^\top \nabla & \mathcal{L}^\text{D}(x', \theta^\text{D}_p, \theta^\text{D}_{r})\}|_{x'\in \mathcal{P}^\text{awy}} \in \mathbb{R}^{L^\text{awy}\times d}, \\
  &G_\text{val}=[G_\text{val}^\text{app}, G_\text{val}^\text{app}] \in \mathbb{R}^{L\times d}.
\end{split}
\end{equation}


\subsection{Data Selection}

In this section, we aim to score the training samples based on the {obtained training sample influence and bidirectional preferences}, 
$G_\text{train} \in \mathbb{R}^{N \times d}$ and $G_\text{val}=[G_\text{val}^\text{app}, G_\text{val}^\text{app}] \in \mathbb{R}^{L\times d}$. 
%
For positive preferences, the correlation between training samples and preferences is determined by their cosine similarity.
%
The correlation matrix $M^\text{app} \in \mathbb{R}^{N\times L^\text{app}}$ can be computed by:
\begin{equation}
    M^\text{app}_{ij}=\frac{{{G_\text{train}}_{i}^\top} \cdot {{G_\text{val}^\text{app}}_{j}}}{\|{{G_\text{train}}_{i}}\| \|{{G_\text{val}^\text{app}}_{j}}\|}.
\end{equation}
%
Subsequently, we calculate the correlation scores between each training sample and the positive preferences by performing a weighted summation using the L2 norm of positive preference direction,
resulting in $\Gamma^\text{app} \in \mathbb{R}^{N}$.
%
%
Following the same procedure, we can ascertain the correlation matrix of the training data with respect to negative preferences, $M^\text{awy} \in \mathbb{R}^{N\times L^\text{awy}}$, and the correlation scores, $\Gamma^\text{awy} \in \mathbb{R}^{N}$.

Afterwards, the training samples are scored by $\Gamma=\Lambda \times \Gamma^\text{app} - (1-\Lambda) \times \Gamma^\text{awy}$, in which $\Lambda \in \mathbb{R}^{N}$ is a instance-level parameter. 
In order to synthesize the correlations with positive and negative preferences intelligently, we use the simulated annealing algorithm 
to automatically optimize $\Lambda$.
The optimization process is shown in Algorithm~\ref{alg:data_selection}, where the objective energy function is:
\begin{equation}
\small
    E(\Lambda) = - \sum_{t=1}^{N}(\Lambda \times \Gamma^\text{app} - (1- \Lambda) \times \Gamma^\text{awy}). 
\end{equation}
%
The energy function aims to encourage high similarity to positive preference while penalizing high similarity to negative preference.
%
%
Based on $\Gamma$, we can obtain a subset of high-quality training samples $\mathcal{\hat{D}_\text{train}}$ to fine-tune LLMs, achieving $\hat{\mathcal{M}}$.
%
Under certain settings, $\hat{\mathcal{M}}$ can even surpass LLMs fine-tuned with the entirety $\mathcal{D}_\text{train}$.

\begin{algorithm}[tb]
   \caption{Automatic Optimization of $\Lambda$}
   \label{alg:data_selection}
\begin{algorithmic}
\small
   \REQUIRE Initial solution $\Lambda_0 \in \mathbb{R}^N$, where each element ${\Lambda_0}_i \sim U(0, 1)$; initial temperature $T_0=1.0$; cooling rate $\alpha=0.95$; termination temperature $T_{\text{end}}=0.01$
\ENSURE Optimal or near-optimal solution $\Lambda$
\STATE $\Lambda \leftarrow \Lambda_0$
\STATE $T \leftarrow T_0$
\WHILE{$T > T_{\text{end}}$}
    \STATE Generate a neighbor solution $\Lambda'$ by perturbing $\Lambda$ with a small random vector $\Delta \Lambda$, where $\Delta \Lambda_i \sim N(0, \sigma)$
    \STATE Calculate the energy difference $\Delta E = E(\Lambda') - E(\Lambda)$
    \IF{$\Delta E < 0$}
        \STATE Accept $\Lambda'$ (i.e., $\Lambda \leftarrow \Lambda'$)
    \ELSE
        \STATE Generate a random number $r \sim U(0, 1)$
        \IF{$r < e^{-\Delta E / T}$}
            \STATE Accept $\Lambda'$ (\textit{i.e.}, $\Lambda \leftarrow \Lambda'$)
        \ENDIF
    \ENDIF
    \STATE $T \leftarrow \alpha T$
\ENDWHILE
\STATE Return the current solution $\Lambda$
\end{algorithmic}
\end{algorithm}

\section{Experiments}
\subsection{Experimental Setup}
\subsubsection{Datasets}

We used the datasets in both target-agnostic and targeted data selection methods. 
%

\textbf{1. Target-agnostic datasets. }
Consistent with the IFD \cite{IFD}, we employed the Alpaca dataset \cite{taori2023stanford}, constructed by Stanford University using the self-instruct \cite{Selfinstruct} paradigm, as the training set. 
This classic high-quality dataset contains 52,002 
triplets. 
%
The corresponding test sets are: 
Vicuna \cite{vicuna}, Koala \cite{Koala}, Wizardlm \cite{WizardLM}, Self-instruct \cite{Selfinstruct}, and LIMA \cite{LIMA}. 
These test sets encompass about 1K human-created open-domain and closed-domain instructions, providing a comprehensive reflection of the model's performance across different scenarios.

\textbf{2. Targeted datasets. }
In line with LESS \cite{LESS}, we used FLAN V2 \cite{longpre2023flan}, COT \cite{wei2022chain}, DOLLY \cite{conover2023free}, and OPEN ASSISTANT \cite{kopf2024openassistant} as training sets, which collectively contain approximately 270K instruction-following examples, covering a variety of data formats and complex reasoning tasks. 
%
For testing, we adopted three widely-used benchmark test sets: MMLU \cite{hendrycks2020measuring}, TYDIQA \cite{clark2020tydi}, and BBH \cite{suzgun2023challenging}.
%
Notably, the training datasets do not include any examples directly related to the target tasks.

%


\subsubsection{Implementation Details}

 In the SFT warm-up phase, we used 5\% of the full training set as $\mathcal{D}_\text{warm}^\text{SFT}$.
 %
 The training data for the DPO warm-up phase is constructed as stated in Sec.~\ref{sec:warm-up}. 
 Only the generated responses judged as different from the ground-truth ones are retained to serve as $r^l$.
 The total number of DPO warm-up pairs is also 5\% of the full training set. 
 %
 %

To assess the similarity between the training data and the target tasks, we construct validation sets for the test sets. 
For Alpaca-related test datasets, we randomly select 10\% of the data as the validation set if there is no sub-tasks.
%
Since Vicuna and Wizardlm test sets contain sub-tasks, we randomly choose 2 samples from each sub-task.
For the targeted datasets, we adopted the same validation set construction method as LESS.
%
More implementation details can be found in Appendix~\ref{app:details}.

\subsubsection{Evaluation Metrics}
For the Alpaca-related test sets, we use pair-wise comparisons to evaluate the instruction-following capabilities, where GPT-4 is used as the comparative evaluator. 
%
%
We calculate the number of ``win'', ``tie'', and ``lose'' and draw the winning score by: $(\#win-\#lose)/\#all+1$. 
%
%
%
%
Building upon LESS \cite{LESS}, we conduct an automatic evaluation for MMLU, TYDIQA, and BBH. 
Please refer to Appendix ~\ref{app:metrix} for more details about the evaluation metrics. 


\subsection{Main Results}


\noindent \textbf{A. Comparison with targeted methods.}

The baseline models include BM25 \cite{bm25}, DSIR \cite{xie2023data}, RDS \cite{hanawa2020evaluation}, and LESS \cite{LESS}, among which LESS is the most related to our method.
%
%
As shown in Tab.~\ref{tab:less_main}, overall, 
larger models outperform smaller ones. 
However, since TYDIQA provides contextual information for answering questions, the generative capabilities of LLMs are less critical, leading to better performance when utilizing the smaller Llama3.2-1B.
Besides, since MMLU contains multiple tasks and the response format is in the form of option numbers, 
the preference contained among different responses is relatively scarce.
Therefore, the improvement of ProDS on MMLU is relatively limited.
Compared to MMLU and TYDIQA, the output of BBH contains CoT and answers, the preferences among diverse responses contain more semantic information. 
Consequently, ProDS achieves greater improvements in BBH.

\begin{table*}
  \centering
  \small
  \caption{Comparison with different targeted data selection methods, where ``1B'' is Llama32-1B and ``7B'' is Llama2-7B for brevity.
  The results for BM25, DSIR, and RDS methods are obtained by training Llama2-7B.
  }
  \begin{tabular}{l|c|ccc|ccccccc}
    \toprule
    \small
    \multirow{2}{*}{\textbf{Dataset}} & {\textbf{Full}}  & {\textbf{BM25} }  & {\textbf{DSIR}}  & {\textbf{RDS}} & \multicolumn{3}{c}{\textbf{LESS} (5\%) } & \multicolumn{3}{c}{\textbf{Ours} (5\%) } & \multirow{2}{*}{\textbf{$\Delta$}} \\
    \cmidrule(lr){6-8} \cmidrule(lr){9-11}
            &   (100\%)    &   (5\%)  &  (5\%)  &    (5\%) & 1B & 1B$\rightarrow$7B & 7B & 1B & 1B$\rightarrow$7B & 7B & \\
    \midrule
    MMLU    & 51.6  & 47.6  &  46.1   &  45.0  &33.4      & 47.2              & \underline{50.2}      &     34.3      &      45.2     &      \textbf{50.8}   & \colorbox{mycolor1}{$\smash\uparrow$0.6}  \\
    TYDIQA  & 54.0    &  52.7  &   44.5  &  46.8   &  53.8    & 49.7              & \underline{56.2}      &    54.7       &    50.7       &     \textbf{56.6 }   & \colorbox{mycolor1}{$\smash\uparrow$0.4} \\
    BBH     & 43.2  &   39.8   &  36.8   &  36.7  &30.4      & \underline{41.9 }             & 41.5      &    33.1       &   {42.9}        &  \textbf{43.1}     &  \colorbox{mycolor1}{$\smash\uparrow$1.2}  \\
    \bottomrule
  \end{tabular}
  \label{tab:less_main}
\end{table*}

\noindent \textbf{B. Comparison with target-agnostic methods.}

\begin{figure}
    \centering
    \includegraphics[width=\linewidth]{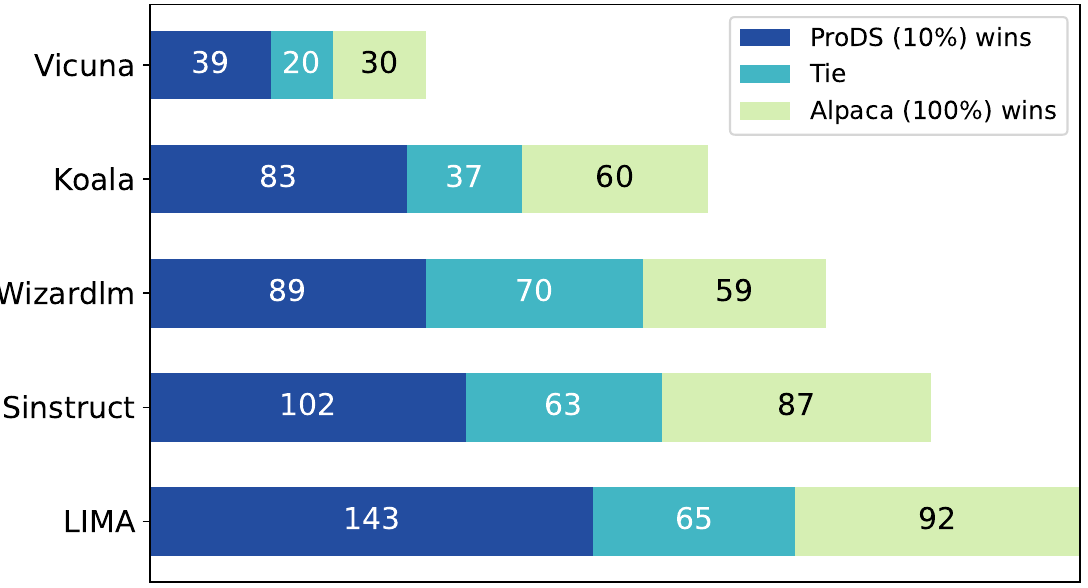}
    \caption{
    Performance of Llama2-7B fine-tuned on the selected top 10\% of Alpaca compared to that fine-tuned on the full Alpaca.
    }
    \label{fig:bars}
\end{figure}

\begin{figure}
    \centering
    \includegraphics[width=\linewidth]{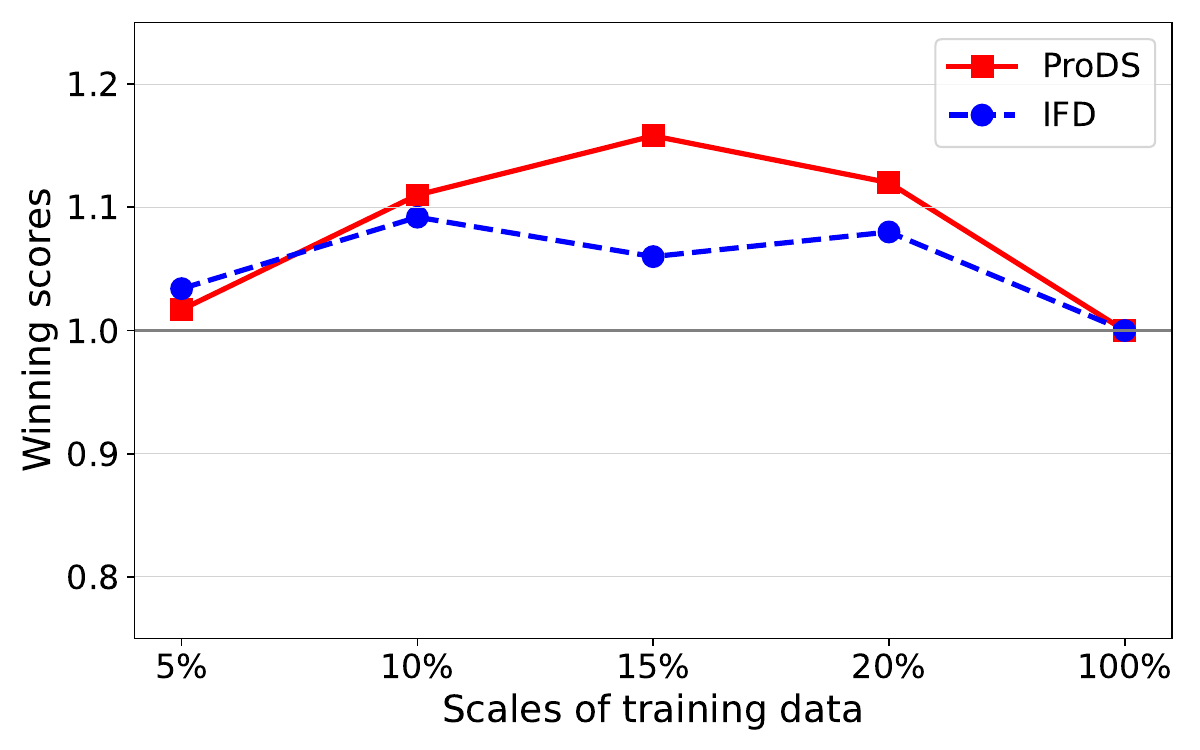}
    \caption{
    Average winning scores of Llama2-7B across the 5 test sets at different data scales.
    } 
    \label{fig:line}
\end{figure}

Fig.~\ref{fig:bars} presents the comparison between 10\% Alpaca chosen by ProDS and full Alpaca on the 5 test sets.
Both the selection LLM and target LLM are Llama2-7B.
%
%
To further validate our instruction selection method, we evaluated the performance under different data scales.
The average winning scores across the 5 test sets are shown in 
Fig.~\ref{fig:line}.
%
Notably, using training sets of varying scales consistently yields win rates greater than 1, indicating that LLMs fine-tuned on our selected data can surpass the full fine-tuning model while reducing the amount of training data.
Besides, our approach also demonstrates advantages with the target-agnostic data selection method, IFD \cite{IFD}. 
This superiority stems from our integration of direct preference learning on top of SFT.
%
%
In contrast, although IFD demonstrates the capability to identify high-quality data through measurement of instruction-following difficulty, it shows limitations in recognizing target-relevant data across medium and lower-quality ranges. 
This explains why IFD marginally outperforms our method at the 5\% selection level, but is consistently surpassed by ProDS as the training data scale increases.

\subsection{Ablation Studies}

\noindent \textbf{A. Different settings to construct $\mathcal{P}^\text{app}$ and $\mathcal{P}^\text{awy}$.}

Our paper utilizes response preferences to select training samples for the specific target.
To construct $\mathcal{P}^\text{app}$ and $\mathcal{P}^\text{awy}$, we utilize two LLMs, $\mathcal{M}_\text{cmp}$ and $\mathcal{M}_\text{base}$ to produce responses of different quality. 
Specifically, the two models are the same LLM trained with different subsets of the training set.
%
To verify whether the choices of $\mathcal{M}_\text{cmp}$ and $\mathcal{M}_\text{base}$ affect the experimental results, we fine-tune three Llama3.2-1B models on different subsets: full Alpaca set (100\%), Top 10\% selected by IFD \cite{IFD},  and randomly selected 10\% from Alpaca.
As shown in Tab.~\ref{tab:dpo_form}, the changes of results are minimal, indicating that our method is robust to the specific DPO construction model.
%

\begin{table}
  \centering
  \small
  \caption{
  Results when using different $\mathcal{M}_\text{cmp}$ and $\mathcal{M}_\text{base}$ to construct validation DPO pairs, 
   where ``Full (100\%)'' denotes full Alpaca set, ``IFD (10\%)'' indicates top 10\% selected by IFD, and ``Rand. (10\%)'' represents a randomly selected 10\% subset of Alpaca.
   %
   All experimental results are evaluated by GPT-4 on the Wizardlm test set.
}
    \begin{tabular}{ccc}
    \toprule
    $\mathcal{M}_\text{base}$ & $\mathcal{M}_\text{cmp}$ & \multicolumn{1}{c}{\textbf{WS} $\uparrow$ [\#w, \#t, \#l]} \\
    \midrule
    Rand. (10\%) & Full (100\%) &  1.03 [73, 79, 66] \\
    IFD (10\%) &  Full (100\%) & 1.05 [72, 87, 59] \\
    Full (100\%) & Rand. (10\%) & \textbf{1.06 [74, 84, 60]} \\
    \bottomrule
    \end{tabular}%
  \label{tab:dpo_form}%
\end{table}%

\noindent \textbf{B. Different methods to compute similarity between validation and training samples.}

Based on $\mathcal{P}^\text{app}$ and $\mathcal{P}^\text{awy}$, we 
compute the similarity scores with positive and negative preferences, and propose an annealing algorithm-based integrating algorithm to synthesize them.
The effectiveness of  data scoring method is validated from the following two perspectives:

\textcircled{1}. \underline{Necessity of separate preference modeling}. 
As a comparison, we constitute a unified set of preference pairs for the outputs from $\mathcal{M}_\text{base}$ and $\mathcal{M}_\text{cmp}$, designating responses with higher GPT-4 scores as $r^w$ and those with lower scores as $r^l$. 
Subsequently, we still compute the preference similarity matrix $M$ between the training and validation sets using cosine similarity, and obtain the training set scores $\Gamma$ by weighted summation of each row in $M$ using the L2 norm of the validation gradients. 
%
As shown in the second row of Tab.~\ref{tab:selection_method}, this unified preference construction approach leads to significant performance degradation. 
This is because the unified preference can only provide the positive optimization direction, 
whereas our separate preference modeling method can additionally offer the negative optimization direction that should be circumvented, resulting in a more comprehensive assessment of the training set's quality.

\textcircled{2}. \underline{Validation of score calculation methods}.
We systematically replace the calculation methods for the intermediate results.
The results are shown in the last part of Tab.~\ref{tab:selection_method}, where ``mul'' represents using multiplication to compute $M^\text{app}$ and $M^\text{awy}$, ``avg'' denotes 
averaging $M^\text{app}$ and $M^\text{awy}$ for $\Gamma^\text{app}$ and $\Gamma^\text{awy}$, and ``fixed'' indicates ``$\Gamma =\Gamma^\text{app} - \Gamma^\text{awy}$''.
It can be seen that there is a certain decrease in the performance after replacing different computation methods.
And modifying the method to synthesize $\Gamma^\text{app}$ and $\Gamma^\text{awy}$ has the most significant impact. 
This proves the effectiveness of our proposed annealing algorithm-based integrating method.

\begin{table}
  \centering
  \caption{Comparison of different data selection methods. ``\#~Pairs'' indicates unifying the validation DPO pairs or using the separate $\mathcal{P}^\text{app}$ and $\mathcal{P}^\text{awy}$.
  Other columns are the methods to obtain intermediate results, 
  $M^\text{xx}$ including $M^\text{app}$ and $M^\text{awy}$, and $\Gamma^\text{xx}$ denoting $\Gamma^\text{app}$ and $\Gamma^\text{awy}$. 
  }
  \resizebox{\linewidth}{!}{
    \begin{tabular}{c|ccc|c}
    \toprule
    \textbf{\# Pairs} & $M^\text{xx}$ & $\Gamma^\text{xx}$ & $\Gamma$ & \textbf{WS}  $\uparrow$ [\#w, \#t, \#l] \\
    \midrule
    Rand. (10\%)     & -   & -   & -    &  0.75 [52, 61, 105]  \\
    \midrule
    unified     & cos   & -   & weight     &   0.83 [54, 72, 92]
\\
    \midrule
    {\multirow{4}[2]{*}{\textbf{separate}}} & \textbf{cos} & \textbf{weight} & \textbf{annealing} & \textbf{1.06 [74, 84, 60]} \\
    & \underline{mul}   & weight   & annealing    & 0.99 [81, 54, 83] \\
  & cos   & \underline{avg}   & annealing    &  	0.95 [66, 75, 77] \\
  & cos   & weight   & \underline{fixed}    & 0.94 [61, 82, 75] \\
          
    \bottomrule
    \end{tabular}%
    }
  \label{tab:selection_method}%
\end{table}%

\noindent \textbf{C. Correlation of selection and target models.}

Since our data selection method is based on gradients of LLMs, we conduct experiments 
to verify whether there exists a potential connection between the selection model used for gradient computation and the target fine-tuning model.
%
We employ Llama3.2-1B \cite{llama3} and Llama2-7B \cite{LLaMa2} as selection models.
The target model covers multiple model architectures including Llama, Qwen \cite{qwen}, Mistral \cite{mistral}, and DeepSeek \cite{bi2024deepseek}. 
The experimental results are winning scores compared with the same target model fine-tuned on full Alpaca.
As shown in Tab.~\ref{tab:backbones}, when keeping the selection model constant, the scale of target models is positively correlated with the final performance. 
When the scale of the target model is the same, using a selection model with consistent architecture yields slightly better results. 
This finding provides an important insight: 
smaller-scale models can be employed for data selection, thereby significantly reducing the computational costs of data selection.

\begin{table}
  \centering
  \small
  \caption{Correlation of data selection models and the target models.
  The target models are fine-tuned on top 10\% of Alpaca selected by the corresponding selection models. 
  The results are the winning scores evaluated on the Wizardlm test set. 
  }
    \begin{tabular}{clc}
    \toprule
    \multicolumn{1}{l}{\textbf{Selection}} & \textbf{Target} & \multicolumn{1}{l}{\textbf{WS} $\uparrow$ [\#w, \#t, \#l]} \\
    \midrule
    \multirow{2}[2]{*}{Llama3.2-1B} & Llama3.2-1B & 1.06 [74, 84, 60] \\
          & Llama2-7B & \textbf{1.12 [92, 62, 64]}  \\
    \midrule
    \multirow{3}[2]{*}{Llama2-7B} & Llama2-7B &   		\textbf{1.14 [89, 70, 59]}  \\
          & Qwen2-7B & 1.12 [89, 67, 62] \\
          & Mistral-7B & 1.03 [78, 68, 72]   \\
          & DeepSeek-7B & 1.02  [78, 67, 73] \\
    \bottomrule
    \end{tabular}%
  \label{tab:backbones}%
\end{table}%

\subsection{Discussion about Selected Data}


\noindent \textbf{A. Instruction distribution. }

\begin{figure}
    \centering
    \includegraphics[width=0.83\linewidth]{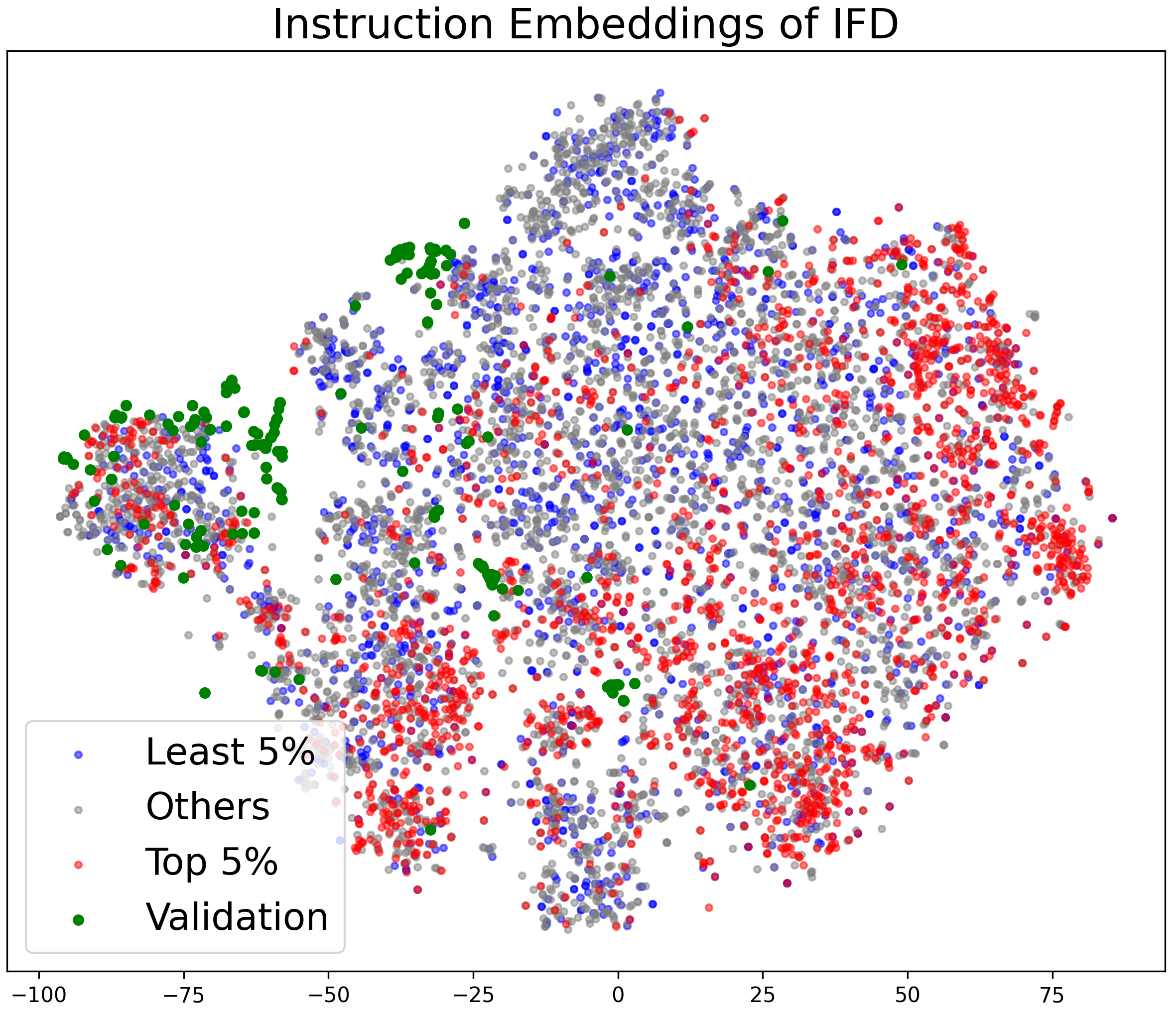}
    \caption{
    Instruction embeddings of top 5\%, least 5\%, and other samples selected by IFD. 
    %
    }
    \label{fig:IFD-embedding}
\end{figure}

\begin{figure}
    \centering
    \includegraphics[width=0.83\linewidth]{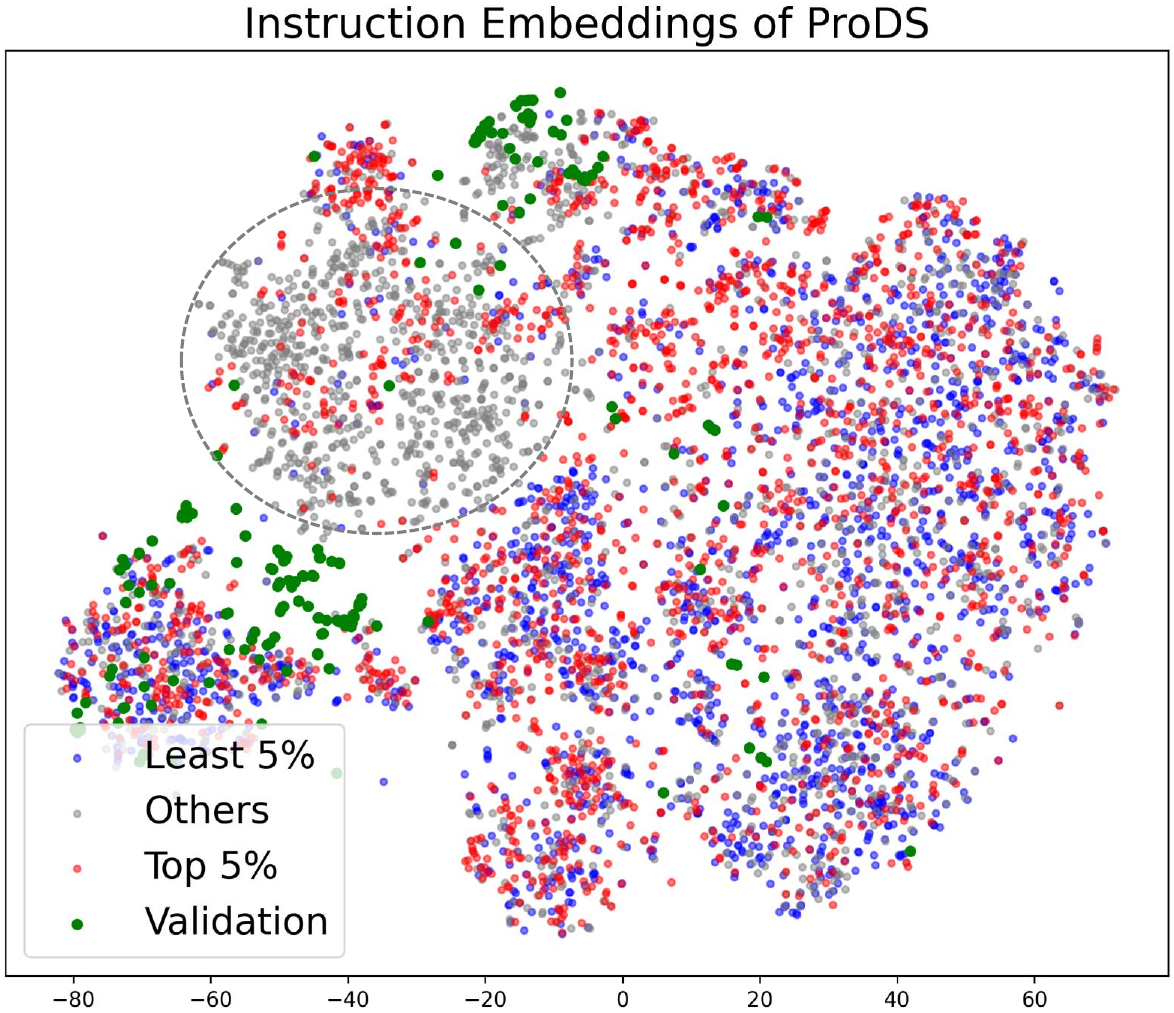}
    \caption{
    Instruction embeddings of top 5\%, least 5\%, and other samples selected by ProDS. 
    %
    %
    }
    \label{fig:Ours-embedding}
\end{figure}

We employ BERT \cite{DBLP:conf/naacl/DevlinCLT19} to compute embeddings of the instructions in the Alpaca dataset, and utilize t-SNE to map them to 2D features. 
We visualize the distribution of top 5\%, least 5\%, and other training samples selected by IFD in Fig.~\ref{fig:IFD-embedding}. The instruction distribution of ProDS is shown in Fig.~\ref{fig:Ours-embedding}. 
The green dots represent the instruction embeddings from the Wizardlm test set.

As illustrated in Fig.~\ref{fig:IFD-embedding}, the IFD method, which directly computes the instruction-following difficulty of training data, demonstrates a clear distinction between its selected top 5\% and least 5\% data. 
However, since the IFD is target-agnostic, the distribution of its top 5\% data exhibits a significant deviation from that of the Wizardlm test set, with an average Euclidean distance of 85.85 after t-SNE reduction. 
%
In contrast, as illustrated in Fig.~\ref{fig:Ours-embedding}, by taking into account the relationship between the target and training sets, our selected top 5\% Alpaca aligns more closely with the distribution of the Wizardlm test set (with an average Euclidean distance of 67.13 after t-SNE reduction). Meanwhile, the selected least 5\% are farther away from Wizardlm.
%
Furthermore, Fig.~\ref{fig:Ours-embedding} also reveals that some of the training samples between the top 5\% and least 5\% (\textit{i.e.}, the gray dots within the gray dashed circle), are also relatively close to the Wizardlm test set, which can explain the superiority of ProDS at 10\%, 15\%, and 20\% in Fig.\ref{fig:line}.

\noindent \textbf{B. Preference of Responses. }

\begin{figure}
    \centering
    \includegraphics[width=0.95\linewidth]{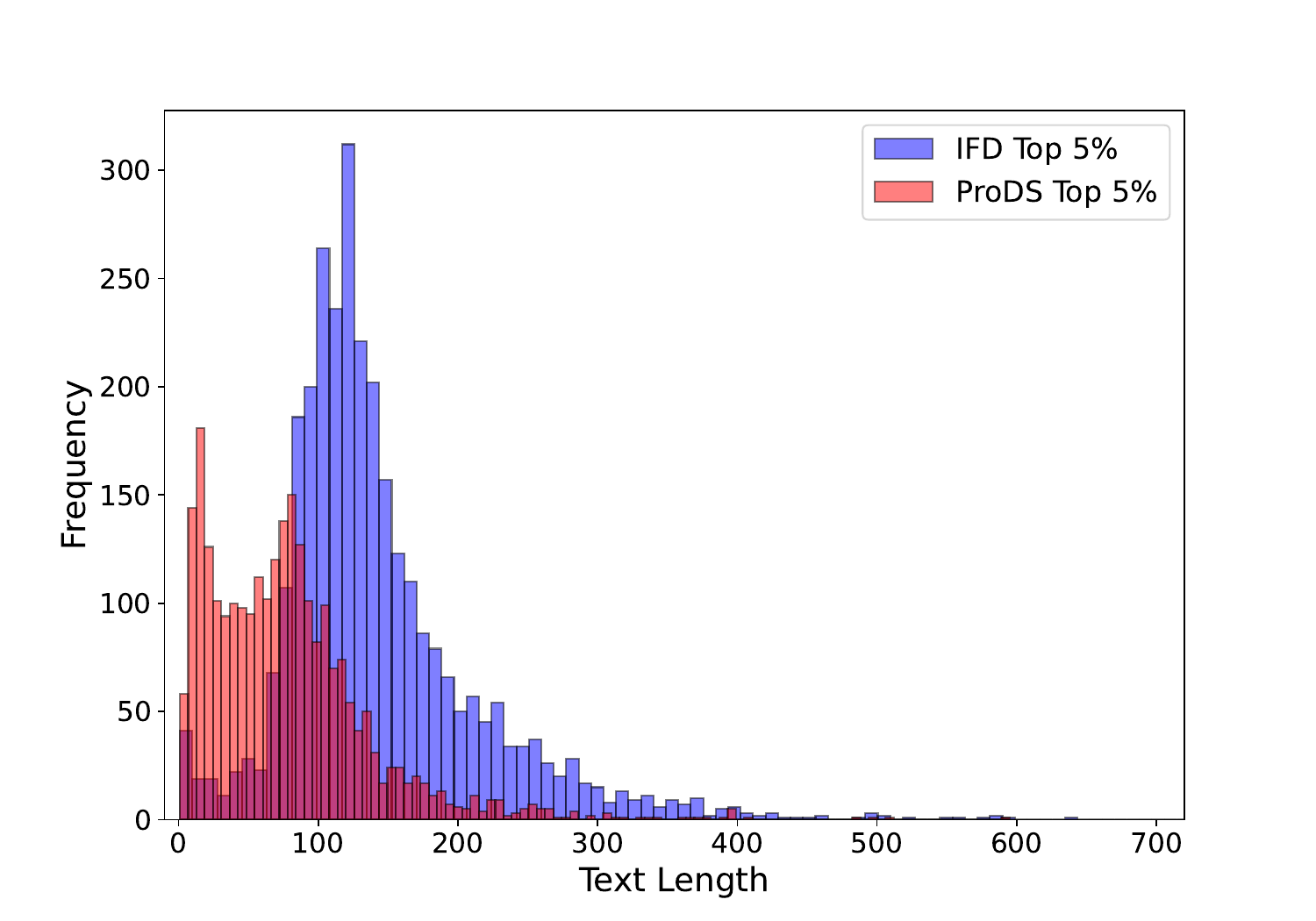}
    \caption{
    Distribution of response lengths for the top 5\% of Alpaca selected by IFD and ProDS.
    %
    }
    \label{fig:len-distribution}
\end{figure}

Our main idea is to select training samples that align with the preferences of target sets.
In this section, we use response length as an intuitive preference indicator to analyze the consistency between the selected data and the Wizardlm test set. 
In addition to the constructed validation set, we also use $\mathcal{M}_\text{cmp}$ and $\mathcal{M}_\text{base}$ to construct DPO pairs for full Wizardlm. 
%
Using GPT-4 as the evaluator, we conduct comparative analysis of response length differentials between preferred and dispreferred outputs. 
The results demonstrates a 7\% reduction in response length for preferred responses, indicating that shorter responses are preferred for this target set. 
%
%
The response lengths of top 5\% data selected by IFD and ProDS are illustrated in Fig.\ref{fig:len-distribution}.
It can be seen that responses in our selected training data are shorter than those selected by IFD, meaning our method can better align with the preferences of the Wizardlm test set in terms of response length.

Besides, to further explore whether response length is the only preference modeled by ProDS, we reconstruct the selected subset. 
While maintaining the semantic distribution remains unchanged, we construct a new 5\% Alpaca set with data that includes responses as long as possible, which deviates most from the length preference of Wizardlm.
After fine-tuning LLM with this new set, we observe minimal changes in performance, indicating that our data selection method can also model other effective response preferences than response length. 
For more details of this exploratory experiment, please refer to Appendix \ref{sec:preference}.

\section{Conclusion}
In this paper, we introduce ProDS, a novel data selection method that can align with the response preference in the target task. 
ProDS utilizes the DPO gradient to estimate target preferences. 
Moreover, we introduce a bidirectional preference synthesis strategy to score the training samples by considering both positive and negative preference directions.
Experimental results compared to existing task-agnostic and targeted methods demonstrate the efficacy of our proposed approach.

\newpage
\section*{Limitations}
Although extensive experiments have confirmed the superiority of ProDS, two major limitations remain.
First, the preference alignment process requires constructing DPO pairs and performing DPO warm-up, which introduces additional time overhead. 
However, it is worth noting that the models used for DPO pair construction are the same SFT warm-up models and baseline models, meaning no additional models are needed. 
Furthermore, this extra computation occurs during offline data selection and does not affect inference efficiency.
Second, the preference modeling in this work relies solely on DPO gradients. 
Exploring more effective preference modeling strategies remains a valuable direction for future research.
%

\bibliography{custom}

\clearpage
\appendix

\begin{table*}
  \centering
  \caption{Statistics of the constructed validation set. }
  \resizebox{0.6\linewidth}{!}{
    \begin{tabular}{lccccc}
    \toprule
    Dataset & \# Shot & \# Task & \multicolumn{1}{c}{$|$ Validation Set $|$} & \multicolumn{1}{c}{$|$ Test Set $|$} & Answer Type \\
    \midrule
    \multicolumn{6}{l}{\textit{LESS-related validation sets}} \\
    \midrule
    MMLU  & 5 & 57 & 285   & 18,721 & Letter options \\
    TYDIQA & 1 & 9 & 9     & 1,713 & Span \\
    BBH   & 3 & 23 & 69    & 920   & COT and answer \\
    \midrule
    \multicolumn{6}{l}{\textit{Alpaca-related validation sets}} \\
    \midrule
    Vicuna & 2 & 9 & 18     & 80    & Answer \\
    Koala & -     & -     & 18     & 180   & Answer \\
    Wizardlm & 2 & 18 & 36    & 218   & Answer \\
    Sinstruct & -     & -     & 25     & 252   & Answer \\
    LIMA  & -     & -     & 30     & 300   & Answer \\
    \bottomrule
    \end{tabular}%
    }
  \label{tab:validation_statis}%
\end{table*}%

\section{Implementation Details.}
\label{app:details}

\subsection{Details about Warm-up}
During the SFT warm-up process, we utilize 5\% of the training set to fine-tune $\mathcal{M}_0$, with a total of 4 training epochs. In the DPO warm-up phase, we initially construct DPO pairs using the training set. Specifically, we leverage the LLM that has been fine-tuned through the SFT warm-up process to generate responses for the training data. Subsequently, GPT-4 is employed to assess the consistency of these generated responses with the ground-truth responses.
Samples deemed consistent with the ground-truth are excluded. 
For the remaining samples, the model-generated responses are designated as $r^l$ (dispreferred responses), while the ground-truth responses in the dataset are labeled as $r^w$ (preferred responses), thus forming DPO pairs. Ultimately, we retain 5\% of the total training data for the DPO warm-up process, with a total of 1 training epoch. It is worth noting that both the SFT and DPO warm-up phases utilize the Adam optimizer for model optimization.
 Since the warm-up and selection paradigm is similar to LESS, we conducted experiments based on the basic framework of LESS and retained most of its training details.

\subsection{Statistics of Validation Datasets}
\label{app:valid_statistics}

For the targeted datasets, MMLU, TYDIQA, and BBH are used as test sets. 
MMLU encompasses a diverse range of 57 multiple-choice tasks, spanning subjects such as elementary mathematics, computer science, etc. 
TYDIQA presents a multilingual question-answering dataset featuring 9 languages, where the challenge lies in extracting answers from provided passages in response to questions. 
Meanwhile, BBH comprises a curated selection of 27 challenging tasks from BIG-Bench, specifically designed to evaluate reasoning abilities.  
Correspondingly, the used validation sets are constructed based on the sub-tasks. 
For the 57 tasks in the MMLU dataset, each task provides 5-shot samples; for the 9 tasks in TYDIQA, each task provides 1-shot samples; and for the 23 datasets in BBH, each dataset provides 3-shot samples. 
%
These details of validation construction process are consistent with LESS.

Among the Alpaca-related test datasets, Vicuna and Wizardlm contain 9 and 18 sub-tasks, respectively. 
We select 2-shots for each sub-task. 
For the other 3 test sets, we randomly select 10\% of the data as the validation set.
%
The detailed statistics about validation sets are shown in Tab.~\ref{tab:validation_statis}.

\subsection{Validation DPO Pair Construction. }
Due to the absence of annotated responses in the Alpaca-related validation set, we implemented a dual-generation mechanism: employing two LLMs fine-tuned on training sets of varying scales to concurrently generate responses. 
With GPT-4 as the evaluator to evaluate the quality of the responses, we construct $\mathcal{P}^\text{app}$ and $\mathcal{P}^\text{awy}$ as stated in the section of methodology. 

Since the validation set used by the targeted models (\textit{i.e.}, MMLU, BBH, and TYDIQA) contain ground-truth responses, we directly utilized the annotated responses to construct DPO pairs as aforementioned details about training DPO pair construction.
However, when processing datasets like MMLU that only annotate the correct answer option letters, we adopted an enhancement strategy: using the complete content of the ground-truth option as $r^w$ while randomly selecting another option as $r^l$, thereby enriching the semantic information contained within the DPO pairs.
For the other two datasets, BBH and TYDIQA, we utilize the warm-up model to generate responses for the instructions. By comparing the generated responses with those in the test set, the incorrect responses are retained to serve as $r^l$ in the DPO pairs.

\subsection{Computational Complexity}
The most time-consuming steps in ProDS are warm-up and gradient computation. 
We conducted the warm-up on 4 A6000 GPUs (48G each).
The SFT warm-up based on 5\% of the Alpaca dataset took a total of 0.5 hours.
The DPO warm-up based on 5\% of the Alpaca dataset took a total of 0.4 hours.
For FLAN V2, COT, DOLLY, and OPEN ASSISTANT used in targeted methods, we also performed warm-up using 5\% of the total training data.
The SFT warm-up and DPO warm-up took 2 hours and 1.5 hours, respectively.
We performed gradient computation on a single A100 GPU (80G).
Gradient computation for the Alpaca dataset took a total of 4 GPU hours.
Gradient computation for the FLAN V2, COT, DOLLY, and OPEN ASSISTANT training sets took a total of 20 GPU hours.
Although these computations are time-intensive, they are one-time efforts for the training sets and can be applied to various target tasks.

\section{Evaluation Metrics}
\label{app:metrix}

For the dataset for target-agnostic methods, we use pairwise evaluation methods. 
Specifically, for each instruction in the test dataset and the corresponding response generated by the models to be compared, we ask GPT-4 to score each model's response on a scale of 1 to 10. 
The response obtaining higher scores is considered as the winner. 
It is noteworthy that, to mitigate potential bias in GPT's responses due to answer ordering, we construct two GPT requests with reversed orders for each instruction: <Response A, Response B> and <Response B, Response A>. 
A response is considered to have won on a given instruction only if it does not fail in both orders and achieves at least one victory across the two orders. 
If the model wins in one order and loses in the other, it is considered a tie. 
Finally, the winning score is calculated by $(\#win-\#lose)/\#all+1$.

For the test sets used in targeted methods, we use the same evaluation methods as LESS. 
For MMLU, we conducted an analysis of the test sets encompassing 57 sub-tasks within the MMLU dataset, subsequently calculating the average 5-shot accuracy for these sub-tasks.
For the TYDIQA dataset, we adopted a 1-shot evaluation methodology, wherein we computed the macro-average F1 score across eleven languages. 
This evaluation is conducted using the gold passage setup, which furnished the model with reference passages containing the standard answers.
Regarding the BBH dataset, we reported the average 3-shot exact match score for all tasks included therein.

\section{Experimental Results about Preference Modeling.}
This experiment aims to verify whether the effectiveness of our method stems solely from alignment with length preferences.
Given that Wizardlm comprises 18 distinct categories, representing information spanning various semantic dimensions, we utilized GPT-3.5 to classify the training samples, so as to differentiate the relevance of the target set and training set in terms of semantic categories.
We collect the distribution of different categories for the top 5\% of data selected by our method, and the full Alpaca. As shown in Fig.~\ref{fig:len-distribution}, the two distributions are largely in agreement, indicating that our selected data preserves the diversity and balance of the original dataset.
%
%
To ablate the influence of length preference, we constructed a new 5\% training subset by selecting the longest responses from the training set while maintaining the same distribution across the 18 categories. 
This subset exhibits a response length distribution opposite to that of the Wizardlm test set. We then fine-tuned the Llama3.2-1B model using this reconstructed data.
The results before and after reconstruction are shown in Tab.~\ref{tab:change_length}. The minimal performance gap suggests that our method captures not only length preferences but also some other semantic preferences. 
Quantifying and analyzing other types of preferences remains a challenging problem, which we leave as part of our future work.
%


\label{sec:preference}

\begin{figure*}
    \centering
    \includegraphics[width=\linewidth]{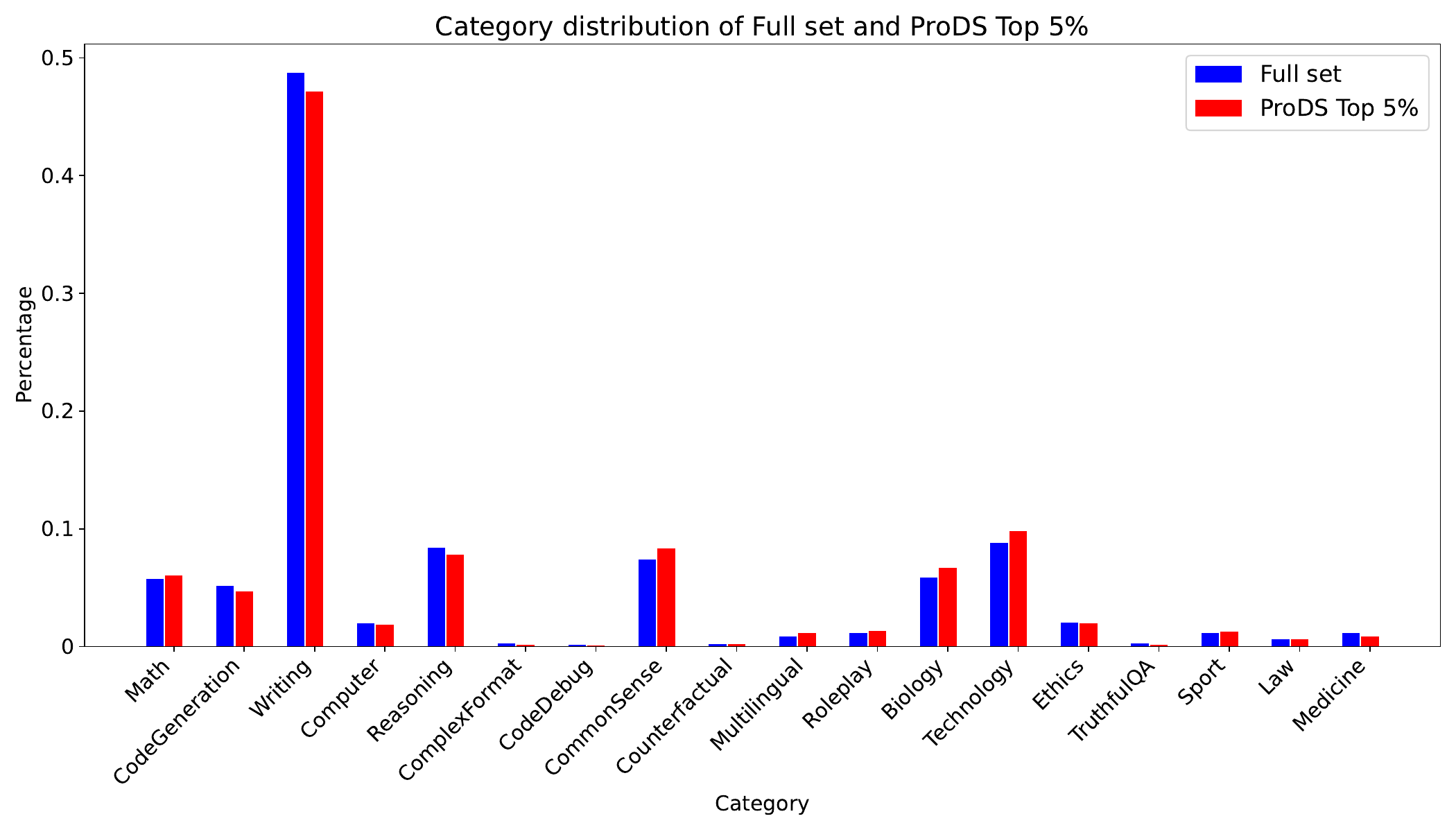}
    \caption{
    Distribution of the top 5\% Alpaca selected by our ProDS and full Alpaca across the 18 categories.
    }
    \label{fig:class_dis}
\end{figure*}


\begin{table}[htbp]
  \centering
  \small
  \caption{Winning scores of 5\% Alpaca before and after the reconstruction process. }
        \begin{tabular}{l|cc}
    \toprule
        Skills  & ProDS 5\% & Resampled 5\% \\
    \midrule
     Sub-Categories &  0.2 / 0.6 / 0.2 &  0.3 / 0.6 / 0.1 \\
     Code Generation &  0.2 / 0.5 / 0.3 &  0.4 / 0.4 / 0.2 \\
     Writting &  0.2 / 0.3 / 0.5 &  0.5 / 0.2 / 0.3 \\
     Computer Science &  0.3 / 0.2 / 0.5 &  0.3 / 0.3 / 0.4 \\
     Reasoning &  0.4 / 0.4 / 0.2 &  0.3 / 0.5 / 0.2 \\
     Complex Format &  0.4 / 0.5 / 0.1 &  0.4 / 0.4 / 0.2 \\
     Code Debug &  0.3 / 0.6 / 0.1 &  0.3 / 0.6 / 0.1 \\
     Common-Sense &  0.5 / 0.4 / 0.1 &  0.4 / 0.6 / 0.0 \\
     Counterfactual &  0.1 / 0.8 / 0.1 &  0.6 / 0.0 / 0.4 \\
     Multilingual &  0.2 / 0.4 / 0.4 &  0.1 / 0.3 / 0.6 \\
     Roleplay &  0.3 / 0.2 / 0.5 &  0.0 / 0.2 / 0.8 \\
     Biology &  0.5 / 0.0 / 0.5 &  0.3 / 0.2 / 0.5 \\
     Technology &  0.1 / 0.2 / 0.7 &  0.3 / 0.2 / 0.5 \\
     Ethics &  0.3 / 0.5 / 0.2 &  0.3 / 0.0 / 0.7 \\
     TruthfulQA &  0.8 / 0.0 / 0.2 &  0.2 / 0.4 / 0.4 \\
     Sport &  0.8 / 0.2 / 0.0 &  0.2 / 0.2 / 0.6 \\
     Law  &  0.4 / 0.2 / 0.4 &  0.0 / 0.6 / 0.4 \\
     Medicine &  0.2 / 0.2 / 0.6 &  0.2 / 0.2 / 0.6 \\
    \midrule
    Wining Score & 1.056  & 1.028 \\
    \bottomrule
    \end{tabular}%
  \label{tab:change_length}%
\end{table}%

\section{Typical Selected Samples. }

We present some selected examples. Each example comprises a given preference pair and the corresponding training data that obtained the highest matching score. From the examples, we can see that our model can select the most suitable data from multiple dimensions.
For example, in the field of open domain Q\&A and dialogue, the sample from Wizardlm in Table~\ref{case:wizardlm} focuses on specificity and originality. In contrast, the sample from Koala in Table~\ref{case:koala} focuses on factual accuracy and completeness.
The samples from LIMA in Table~\ref{case:lima} and Vicuna in Table~\ref{case:vicuna} focus on details and structures.
In the field of arithmetic, the sample from MMLU in Table~\ref{case:mmlu}, the sample from TYDIQA in Table~\ref{case:tydiqa}, and the sample from BBH in Table~\ref{case:bbh} all focus on clear and correct explanations and conclusions.

\begin{table*}[htbp]
    \centering
    \caption{An example of selected train sample for the preference in Wizardlm test set. The selected training data demonstrates precise vocabulary enhancement, aligning with the validation preference on specificity and originality. The preferred response in the preference data avoids clichés and provides concrete details, while the rejected response relies on generic phrases.}
    \begin{tabular}{p{0.95\textwidth}} 
    \toprule
    \multicolumn{1}{c}{\textbf{An example of Wizardlm validation preference}}
    \\
    \midrule
    \textbf{[Instruction]:} \\
     I have a hobby of writing short stories in various genres and I want to publish them online on a platform where other writers and readers can interact and give feedback. I want to attract more readers and make them interested in my stories. Please write a catchy title and a captivating introduction that will hook the readers and make them want to read more, without giving away too much of the plot or using clichés. \\
    \textbf{[Preferred Response]:} \\
    The Mystery of the Missing Mansion: A Thrilling Tale of Suspense and Surprise \\ 
    The story begins with a young couple, John and Sarah, who are on the hunt for their dream home. They stumble upon an old mansion, seemingly abandoned, with a mysterious and alluring aura surrounding it. As they explore the grounds, they discover a hidden room filled with ancient artifacts and secrets. As they delve deeper into the mystery, they uncover a dark and sinister plot that has been hidden for centuries.\\ 
    \textbf{[Dispreferred Response]:} \\
    A captivating title for this story could be “The Mysterious Adventures of a Wandering Writer” or “The Unexpected Journey of a Wandering Writer”. The introduction could be something like “A wandering writer embarks on an unforgettable journey of discovery, filled with unexpected twists and turns that will leave you on the edge of your seat”. This introduction could also include a hint of the main conflict, but without giving too much away. \\
    \midrule
    \multicolumn{1}{c}{\textbf{An example of selected training data}} \\
    \midrule
    \textbf{[Instruction]:} \\
    Rewrite the following sentence using a more sophisticated phrase than “very happy”: \\ I am very happy about my new job. \\
    \textbf{[Response]:} \\
    I am ecstatic about my new job. \\
    \bottomrule
    \end{tabular}
    \label{case:wizardlm}
\end{table*}

\begin{table*}
    \centering
    \caption{An example of selected train sample for the preference in Koala test set. The selected training data and the preferred response both prioritizes factual accuracy and completeness, whereas the rejected response lacks comparable depth. Training with the selected data would reinforce the preference for detailed, contextually rich answers.}
    \begin{tabular}{p{0.95\textwidth}} 
    \toprule
    \multicolumn{1}{c}{\textbf{An example of Koala validation preference}} \\
    \midrule
    \textbf{[Instruction]:} \\
    Write description for 925 sterling silver miami cuban link chain \\
    \textbf{[Preferred Response]:} \\
    The 925 sterling silver Miami Cuban link chain is a classic and timeless piece of jewelry. The chain is made up of small, interlocking links that create a unique and stylish look. The chain is available in a variety of lengths, from 16 inches to 24 inches, and is adjustable for a custom fit. The Miami Cuban link chain is crafted from 925 sterling silver, making it a durable and long-lasting piece of jewelry. It is also hypoallergenic, making it ideal for those with sensitive skin. The Miami Cuban link chain is the perfect choice for any occasion, from casual to formal. \\
    \textbf{[Dispreferred Response]:} \\
    The 925 sterling silver Miami Cuban link chain is a classic piece of jewelry that is perfect for any occasion. It features a sleek and modern design with a classic Cuban link pattern. The chain is made of 925 sterling silver, making it durable and long-lasting. It is also adjustable, allowing you to customize the length of the chain to fit your needs. The Miami Cuban link chain is the perfect addition to any wardrobe. \\
    \midrule
    \multicolumn{1}{c}{\textbf{An example of selected training data}} \\
    \midrule
    \textbf{[Instruction]:} \\
    Describe the species X. \\
    \textbf{[Input]:} \\
    Species X \\
    \textbf{[Response]:} \\
    Species X is a species of small, nocturnal mammal native to tropical rain forests. It has a distinctive black and white fur pattern, a long, prehensile tail, and sharp teeth adapted for a carnivorous diet. Its diet consists of small insects, mice, and other small animals. \\
    \bottomrule
    \end{tabular}
    \label{case:koala}
\end{table*}

\begin{table*}
    \centering
    \caption{An example of selected train sample for the preference in LIMA. The  selected train data provides a detailed, structured, and context-rich answer, which could reinforce the preference for thoroughness over brevity, aligning with the preference hierarchy.}
    \begin{tabular}{p{0.95\textwidth}} 
    \toprule
    \multicolumn{1}{c}{\textbf{An example of LIMA validation preference}} \\
    \midrule
    \textbf{[Instruction]:} \\
    What are some interesting periods in African history? Help me find a setting for my historical novel. \\
    \textbf{[Preferred Response]:} \\
    African history is rich and diverse, with a wide range of interesting periods to explore. One period of particular interest is the 18th century, when the transatlantic slave trade was at its height. This period saw the rise of powerful kingdoms such as the Oyo Empire, the Fulani Empire, and the Ashanti Empire. It also saw the rise of Islam in West Africa and the development of a number of cultural and artistic traditions. Another period of interest is the 19th century, when the European powers began to colonize parts of Africa. This period saw the emergence of new nations, the development of new political and economic systems, and the emergence of new forms of resistance. Finally, the 20th century saw the development of modern African nations, the rise of pan-Africanism, and the emergence of new forms of culture and art. \\
    \textbf{[Dispreferred Response]:} \\
    Some interesting periods in African history include the slave trade era, the colonial period, and the independence era. The slave trade era spanned from the 15th to the 19th centuries, during which millions of Africans were forcibly transported to other continents as slaves. The colonial period lasted from the 19th to the mid-20th century, during which African countries were ruled by European powers. The independence era, which began in the mid-20th century, saw the independence of many African countries from European rule. \\
    \midrule
    \multicolumn{1}{c}{\textbf{An example of selected training data}} \\
    \midrule
    \textbf{[Instruction]:} \\
    Write a news article about the event.: The opening ceremony of the Tokyo 2020 Olympics was held at the National Stadium. \\
    \textbf{[Response]:} \\
    The long awaited Tokyo 2020 Olympics officially kicked off today with a grand opening ceremony at the National Stadium. The stadium was packed with spectators from all over the world, ready to catch a glimpse of the spectacular event. Spectacular visuals lit up the night as the athletes entered the stadium carrying the iconic Olympic flame and displaying the flags of their respective countries. \\
    Athletes and spectators alike were mesmerized by the dynamic displays of color and light while favorite tunes played in the background. The night’s program featured performances by top singers, dancers, and other entertainers from around the globe. The energetic crowd gave a heart-warming welcome to each country’s representatives, providing a truly international atmosphere. \\
    The Olympic flame has been officially lit and its journey has begun! With the beautiful opening ceremony now over, we can expect the Tokyo 2020 Olympic Games to be one of the most memorable events in sporting history. \\
    \bottomrule
    \end{tabular}
    \label{case:lima}
\end{table*}

\begin{table*}
    \centering
    \caption{An example of selected train sample for the preference in Vicuna test set. The selected training data emphasizes intensifying emotion, aligning with the preferred response's focus on heightened emotional language.}
    \begin{tabular}{p{0.95\textwidth}} 
    \toprule
    \multicolumn{1}{c}{\textbf{An example of Vicuna validation preference}} \\
    \midrule
    \textbf{[Instruction]:} \\
    You are a mountain climber reaching the summit of Mount Everest. Describe your emotions and the view from the top. \\
    \textbf{[Preferred Response]:} \\
    I feel overwhelmed with excitement and awe as I stand at the summit of Mount Everest. The view is breathtaking, with snow-capped mountains in every direction and a vast blue sky above. \\
    \textbf{[Dispreferred Response]:} \\
    I am filled with a sense of accomplishment and awe as I gaze upon the world below me. The view is breathtaking, with the snow-capped peaks of the Himalayas in the distance and the valleys below dotted with villages and forests. I feel a deep sense of peace and gratitude for the journey I have taken to reach this moment. \\
    \midrule
    \multicolumn{1}{c}{\textbf{An example of selected training data}} \\
    \midrule
    \textbf{[Instruction]:} \\
    Edit the sentence to describe the emotion of the people.: The people in the park were pleased. \\
    
    \textbf{[Response]:} \\
    The people in the park were elated. \\
    \bottomrule
    \end{tabular}
    \label{case:vicuna}
\end{table*}

\begin{table*}[htbp]
    \centering
    \caption{An example of selected train sample for the preference in MMLU. The selected training data demonstrates accurate, step-by-step problem-solving in basic arithmetic, reinforcing clarity and correctness.}
    \begin{tabular}{p{0.95\textwidth}} 
    \toprule
    \multicolumn{1}{c}{\textbf{An example of MMLU validation preference}} \\
    \midrule
    \textbf{[Instruction]:} \\
    The following are multiple choice questions (with answers) about  abstract algebra.\\
    Find all $c$ in $Z_3$ such that $Z_3[x]/(x^2 + c)$ is a field.\\
    A. 0 \quad B. 1 \quad C. 2 \quad D. 3 \quad B\\
    Statement 1 | If aH is an element of a factor group, then |aH| divides |a|. Statement 2 | If H and K are subgroups of G then HK is a subgroup of G.\\
    A. True, True  \quad B. False, False \quad C. True, False \quad D. False, True \quad B\\
    Statement 1 | Every element of a group generates a cyclic subgroup of the group. Statement 2 | The symmetric group $S_10$ has 10 elements.\\
    A. True, True \quad 
    B. False, False \quad 
    C. True, False \quad 
    D. False, True \quad  C\\
    Statement 1| Every function from a finite set onto itself must be one to one. Statement 2 | Every subgroup of an abelian group is abelian.\\
    A. True, True \quad 
    B. False, False \quad 
    C. True, False \quad 
    D. False, True \quad A\\
    Find the characteristic of the ring 2Z.\\
    A. 0 \quad 
    B. 3 \quad 
    C. 12 \quad 
    D. 30  \quad A\\
    How many homomorphisms are there of Z into $Z_2$? \\
    A. 1 \quad 
    B. 2 \quad 
    C. infinitely many \quad 
    D. 0 \\
    \textbf{[Preferred Response]:} \\
    Answer:B \\
    \textbf{[Dispreferred Response]:} \\
    Answer:A \\
    \midrule
    \multicolumn{1}{c}{\textbf{An example of selected training data}} \\
    \midrule
    \textbf{[Instruction]:} \\
    Imagine a question and stream-of-consciousness explanation for which this is the answer: 3600 \\
    \textbf{[Response]:} \\
    Question: A teacher teaches 5 periods a day and works 24 days a month. He is paid \$5 per period. if he has been working for 6 months now how much has he earned in total?\\
    Stream-of-consciousness: The amount paid to the teacher per day is 5 periods * \$5 / period = \$25 per day. The amount paid for 24 days is \$25 / day * 24 days = \$600. The total amount for 6 months is \$600 * 6 = \$3600. \\
    \bottomrule
    \end{tabular}
    \label{case:mmlu}
\end{table*}

\begin{table*}
    \centering
    \caption{An example of selected train sample for the preference in TYDIQA. The selected training data correctly deduces logical entailment with a concise, accurate explanation, mirroring the preferred response's factual precision and relevance.}
    \begin{tabular}{p{0.95\textwidth}} 
    \toprule
    \multicolumn{1}{c}{\textbf{An example of TYDIQA validation preference}} \\
    \midrule
    \textbf{[Instruction]:} \\
    nswer the following question based on the information in the given passage.\\
    \textbf{Passage}: The International Security Assistance Force (ISAF) was a NATO-led security mission in Afghanistan, established by the United Nations Security Council in December 2001 by Resolution 1386, as envisaged by the Bonn Agreement.[1][2] Its main purpose was to train the Afghan National Security Forces (ANSF) and assist Afghanistan in rebuilding key government institutions, but was also engaged in the 2001–present war with the Taliban insurgency.\\
    \textbf{Question}: When was the International Security Assistance Force started?" \\
    \textbf{[Preferred Response]:} \\
    Answer: December 2001 \\
    \textbf{[Dispreferred Response]:} \\
    Answer: Not known \\
    \midrule
    \multicolumn{1}{c}{\textbf{An example of selected training data}} \\
    \midrule
    \textbf{[Instruction]:} \\
    Q: Test for natural language inference.\\
    Premise: "Asian men waiting in line many with earphones in their ears."\\
    Hypothesis: "The asian men are standing together."\\
    Is the hypothesis entailed by the premise?\\
    Options: \\
    - yes \\
    - no \\
    - it is not possible to tell\\
    A: yes\\
    Explain how we arrive at this answer: \\
    \textbf{[Response]:} \\
    Explanation: The men are waiting in line which implies that they are standing together. \\
    \bottomrule
    \end{tabular}
    \label{case:tydiqa}
\end{table*}

\begin{table*}
    \centering
    \caption{An example of selected train sample for the preference in BBH. The selected training data and preferred response both employ structured, step-by-step reasoning to arrive at factually accurate conclusions grounded in domain knowledge.}
    \begin{tabular}{p{0.95\textwidth}} 
    \toprule
    \multicolumn{1}{c}{\textbf{An example of BBH validation preference}} \\
    \midrule
    \textbf{[Instruction]:} \\
    Is the following sentence plausible? "Wayne Rooney watched the pitch go by." \\
    \textbf{[Preferred Response]:} \\
    Let's think step by step. Wayne Rooney is an English soccer player. Watching the pitch go by is not part of soccer. So the answer is no. \\
    \textbf{[Dispreferred Response]:} \\
    Let's think step by step. Wayne Rooney is an English soccer player. Watching the pitch go by is part of soccer. So the answer is yes. \\
    \midrule
    \multicolumn{1}{c}{\textbf{An example of selected training data}} \\
    \midrule
    \textbf{[Instruction]:} \\
    Test for natural language inference.\\
    Premise: "A costume party in full swing."\\
    Hypothesis: "Everyone at the party is in regular clothes."\\
    Is the hypothesis entailed by the premise?\\
    Options:\\
    - yes\\
    - no\\
    - it is not possible to tell Step by step answer: \\
    \textbf{[Response]:} \\
    Everyone is either in a costume or regular clothes at a party. Therefore, the final answer is no. \\
    \bottomrule
    \end{tabular}
    \label{case:bbh}
\end{table*}

\end{document}